\documentclass[sigconf, nonacm, pdfa]{acmart}

\newcommand\vldbdoi{10.14778/3712221.3712222}
\newcommand\vldbpages{516 - 529}
\newcommand\vldbvolume{18}
\newcommand\vldbissue{3}
\newcommand\vldbyear{2024}
\newcommand\vldbauthors{\authors}
\newcommand\vldbtitle{\shorttitle} 
\newcommand\vldbavailabilityurl{https://github.com/qzc438/ontology-llm}
\newcommand\vldbpagestyle{empty} 

\usepackage{url}
\usepackage{enumitem,kantlipsum}
\usepackage{adjustbox}
\usepackage{multirow}
\usepackage{makecell}
\usepackage{listings}
\usepackage{balance}
\usepackage{subfig}
\usepackage{pdfpages}

\usepackage{color}
\definecolor{dkgreen}{rgb}{0,0.6,0}
\definecolor{gray}{rgb}{0.5,0.5,0.5}
\definecolor{mauve}{rgb}{0.58,0,0.82}
\begin{document}
\title{Agent-OM: Leveraging LLM Agents for Ontology Matching}

\author{Zhangcheng Qiang}
\affiliation{
  \institution{Australian National University}
  \city{Canberra}
  \state{ACT}
  \country{Australia}
}
\email{qzc438@gmail.com}

\author{Weiqing Wang}
\affiliation{
  \institution{Monash University}
  \city{Melbourne}
  \state{VIC}
  \country{Australia}
}
\email{teresa.wang@monash.edu}

\author{Kerry Taylor}
\affiliation{
  \institution{Australian National University}
  \city{Canberra}
  \state{ACT}
  \country{Australia}
}
\email{kerry.taylor@anu.edu.au}

\begin{abstract}
Ontology matching (OM) enables semantic interoperability between different ontologies and resolves their conceptual heterogeneity by aligning related entities. OM systems currently have two prevailing design paradigms: conventional knowledge-based expert systems and newer machine learning-based predictive systems. While large language models (LLMs) and LLM agents have revolutionised data engineering and have been applied creatively in many domains, their potential for OM remains underexplored. This study introduces a novel agent-powered LLM-based design paradigm for OM systems. With consideration of several specific challenges in leveraging LLM agents for OM, we propose a generic framework, namely Agent-OM (Agent for Ontology Matching), consisting of two Siamese agents for retrieval and matching, with a set of OM tools. Our framework is implemented in a proof-of-concept system. Evaluations of three Ontology Alignment Evaluation Initiative (OAEI) tracks over state-of-the-art OM systems show that our system can achieve results very close to the long-standing best performance on simple OM tasks and can significantly improve the performance on complex and few-shot OM tasks.
\end{abstract}

\maketitle

\pagestyle{\vldbpagestyle}
\begingroup\small\noindent\raggedright\textbf{PVLDB Reference Format:}\\
\vldbauthors. \vldbtitle. PVLDB, \vldbvolume(\vldbissue): \vldbpages, \vldbyear.\\
\href{https://doi.org/\vldbdoi}{doi:\vldbdoi}
\endgroup
\begingroup
\renewcommand\thefootnote{}\footnote{\noindent
This work is licensed under the Creative Commons BY-NC-ND 4.0 International License. Visit \url{https://creativecommons.org/licenses/by-nc-nd/4.0/} to view a copy of this license. For any use beyond those covered by this license, obtain permission by emailing \href{mailto:info@vldb.org}{info@vldb.org}. Copyright is held by the owner/author(s). Publication rights licensed to the VLDB Endowment. \\
\raggedright Proceedings of the VLDB Endowment, Vol. \vldbvolume, No. \vldbissue\
ISSN 2150-8097. \\
\href{https://doi.org/\vldbdoi}{doi:\vldbdoi} \\
}\addtocounter{footnote}{-1}\endgroup

\ifdefempty{\vldbavailabilityurl}{}{
\vspace{.3cm}
\begingroup\small\noindent\raggedright\textbf{PVLDB Artifact Availability:}\\
The source code, data, and/or other artifacts have been made available at \url{\vldbavailabilityurl}.
\endgroup
}

\section{Introduction}
\label{sec: introduction}

Large language models (LLMs) are pre-trained with an enormous corpus of common knowledge and therefore have powerful generative capabilities. Despite the success of using LLMs in a wide range of applications, leveraging LLMs for downstream tasks still has several challenges. (1) LLMs are pre-trained models that do not capture late-breaking information. (2) LLM hallucinations are often observed in domain-specific tasks and hamper their reliability. LLMs often generate unsound responses that are syntactically sound but factually incorrect~\cite{ji2023survey}. (3) LLMs are good models of linguistic competence, but have shown limited capabilities in non-linguistic tasks, such as planning and routing. LLMs were originally designed for sequential question answering (QA), but most real-world tasks are designed with complex logic and do not follow a straightforward single path~\cite{valmeekam2023large}.

To overcome the limitations of LLM customisation for downstream tasks, LLM-based autonomous agents have become a prominent research area. In the field of artificial intelligence (AI), the notion of agents was first introduced in the famous Turing Test~\cite{turing2009}, referring to intelligent computational entities that can display human-like behaviours. Such AI agents have fallen short of human-level capabilities, as they can only act on simple and heuristic policy functions learnt from constrained environments and they lack efficient central control to simulate the human learning process~\cite{wang2023survey}. LLMs, with remarkable success in demonstrating autonomy, reactivity, pro-activeness, and social ability, have attracted growing research efforts aiming to construct AI agents, so-called LLM agents~\cite{xi2023rise}.

The core concept of LLM agents is to employ the LLM as a controller rather than as a predictive model only (a.k.a. Model as a Service). LLM agents extend LLM capabilities with advanced planning, memory, and pluggable tools, and allow LLMs to communicate with open-world knowledge~\cite{llm-agent}. (1) Planning breaks down a complex task into simpler and more manageable subtasks. LLMs can also receive feedback on plans and perform reflection and refinement. The most practical technique used for LLM planning is chain-of-thought (CoT)~\cite{wei2023chainofthought}. (2) Tools allow LLMs to call external resources for additional information. They are often invoked by LLM actions. (3) Memory provides context to inherently stateless LLMs, including short-term memory and long-term memory. Short-term memory can be considered as context information obtained from planning and tools via in-context learning (ICL)~\cite{brown2020language}. Long-term memory often uses database storage with retrieval-augmented generation (RAG)~\cite{lewis2021retrievalaugmented} to retain information. Unlike fine-tuning, where models need to be retrained to learn new context data, ICL/RAG instead augments the LLM prompts with new information. ICL/RAG is more scalable for working with dynamic information. Almost 90\% of use cases can be achieved by ICL/RAG-based search and retrieval~\cite{gptrag}. A recent paper~\cite{ovadia2023finetuning} demonstrates that RAG surpasses fine-tuning across a diversity of knowledge-intensive tasks.

Ontology matching (OM) is a classic alignment task, aiming to find possible correspondences between a pair of ontologies~\cite{euzenat2007ontology}. OM systems are developed to automate the matching process. There are two dominant design paradigms for OM systems: traditional knowledge-based OM systems that implement pre-defined logic and expert knowledge; and more recent learning-based OM systems that transform the matching task into machine-enhanced learning and prediction. The former expert systems require extensive expert knowledge, while the latter predictive systems need extensive high-quality data to train the model. The prevalence of LLMs and LLM agents has driven many successful domain-specific applications. However, in the context of OM, using LLMs and LLM agents is currently under-explored. Leveraging LLMs and LLM agents for OM tasks is not an intuitive task; the challenges will be presented in the Related Work section.

This paper introduces a novel agent-powered LLM-based design paradigm for OM systems. We propose a generic framework and implement it with a proof-of-concept system. The system extends LLM capabilities beyond general QA, offering a powerful problem solver for OM tasks. The system includes tools that facilitate information retrieval, entity matching, and memory storage. The system is compared to state-of-the-art OM systems, achieving considerable matching performance improvements across three Alignment Evaluation Initiative (OAEI)~\cite{oaei} tracks. Specifically, this paper makes the following contributions:

\begin{itemize}[wide, noitemsep, topsep=0pt, labelindent=0pt]

\item We introduce a new agent-powered LLM-based design paradigm for OM systems and propose a novel Agent-OM framework. It consists of the following key components:
\begin{itemize}[wide, noitemsep, topsep=0pt, labelindent=0pt]
\item A LLM acts as a central ``brain'' to link different modules and instruct their functions via prompt engineering;
\item A pair of planning modules use CoT for OM decomposition;
\item A set of OM tools use ICL/RAG to mitigate LLM hallucinations;
\item A shared memory module uses dialogue and hybrid data storage to support the search and retrieval of entity mappings.
\end{itemize}

\item We implement our proposed Agent-OM framework in a proof-of-concept system. The system deals with several critical downstream challenges in leveraging LLM agents for OM, such as cost-effective entity information retrieval, matching candidate selection, and search-based matching functions.

\item The experimental results of the system show that Agent-OM achieves results very close to the best long-standing performance on simple OM tasks and significantly improves matching performance on complex and few-shot OM tasks.

\end{itemize}

An ontology contains classes, properties, and individuals. In this study, we consider only classes and properties, and individuals are excluded. Possible logical relations between classes (respectively properties) can be equivalence ($\equiv$) and subsumption (either $\subseteq or \supseteq$). In this study, we only consider the logical relation of equivalence ($\equiv$) between classes and properties.

The rest of the paper is organised as follows. Section~\ref{sec: related work} reviews related work. We illustrate the design of our agent-powered LLM-based OM framework in Section~\ref{sec: approach} and present implementation details in Section~\ref{sec: implementation}. Section~\ref{sec: evaluation} and~\ref{sec: ablation} evaluate the system, with a discussion in Section~\ref{sec: discussion}. We discuss the limitations and future work in Section~\ref{sec: limitations} and~\ref{sec: future work}, respectively. Section~\ref{sec: conclusion} concludes the paper.

\section{Related Work}
\label{sec: related work}

OM is typically a non-trivial but essential alignment task for data integration, information sharing, and knowledge discovery~\cite{shvaiko2011ontology}. While matching is a prerequisite for interoperating applications with heterogeneous ontologies, OM systems are designed to automate the matching process. The conventional approach using knowledge-based OM systems, such as LogMap~\cite{jimenez2011logmap1,jimenez2011logmap2}, AgreementMakerLight (AML)~\cite{faria2013agreementmakerlight,faria2014agreementmakerlight}, and FCA-Map~\cite{zhao2018matching,li2021combining} has been shown to be precise and effective. However, it is resource-hungry and labour-intensive. It is often difficult to find domain experts to evaluate the matches, and any group of experts may not be able to cover all domain concepts that an expert system requires. A new approach uses machine learning (ML), implemented in systems such as BERTMap~\cite{he2022bertmap} and LogMap extension LogMap-ML~\cite{chen2021augmenting}. ML-based OM systems employ the concept of training and testing in ML, using ontology entities as features for model training or fine-tuning and then using the model to predict additional correspondences. Specifically, the leading system BERTMap uses a common language model originating from natural language processing (NLP) (i.e. BERT~\cite{devlin2019bert}).

Although ML-based approaches have shown a significant improvement in matching performance, their training-testing paradigm is not feasible for LLMs. The number of parameters used in LLMs is much larger than those of common language models. This means retraining the entire LLM is usually infeasible, and fine-tuning such a large model requires a number of samples that may be infeasibly large for OM. A survey in~\cite{zong2022survey} implies that 1000 is a reasonable number of training samples to fine-tune GPT-3~\cite{brown2020language}, but generally speaking, a domain ontology has only around 100-200 entities. Furthermore, currently some LLMs are only accessible through a web service. This means that training or fine-tuning LLMs risks leaking sensitive information, while synthetic data, on the other hand, makes it difficult to ensure training quality.

Early studies using LLMs for OM can be found in~\cite {he2023exploring} and~\cite{norouzi2023conversational}. Both works use a purely prompt-based approach. The prompts are structured as a binary question: given an entity from the source ontology and an entity from the target ontology, the LLMs perform a classification task to determine whether these two entities are identical or not. A similar approach is also used in OLaLa~\cite{hertling2023olala} and LLMs4OM~\cite{giglou2024llms4om}, but their candidate generation is integrated with the embedding extractor models. The authors of~\cite{amini2024towards} explore the potential of using LLMs for complex ontology OM challenges.

LLM agents were introduced in AutoGPT~\cite{autogpt} and BabyAGI~\cite{babyagi}. The recent release of OpenAI GPTs~\cite{gpts}, Microsoft Copilot~\cite{copilot}, and Copilot Studio~\cite{copilotstudio} has sparked interest in LLM agents. Building applications with LLM agents allows users to build their own custom GPTs to support custom business scenarios. In ontology-related tasks, LLM-driven agents have shown impressive performance in automating manual activities in the broader task of ontology engineering. These works pay attention to the use of conversational dialogue to enhance the agent's capabilities with human feedback. While this is suitable for tasks that require humans to be in the loop, such as collecting competency questions in ontology engineering~\cite{zhang2024ontochat} or validating extended terms in ontology learning~\cite{giglou2023llms4ol}, modern OM seeks to automate a complex task with minimal human intervention. Contrasting with these works, our aim is to design a new infrastructure that is able to instruct LLM agents to use planning to decompose a complex task into steps and to use tools to facilitate automated matching (a.k.a. function calling), instead of purely using agent-based conversational dialogue, even when specialised as ontology-oriented dialogue like~\cite{payne2014negotiating,zhang2024large}.

We introduce our novel agent-powered LLM-based design paradigm for OM systems. We have two generic agents; each one is self-contained and designed to instruct LLMs to use extensive planning, memory, and tools, thus unlocking their generative capabilities to handle various types of OM tasks in different contexts. Meanwhile, as a key enabler for precise decision-making, we also limit the current LLM's flaws in hallucination, context understanding, and non-linguistic reasoning. Several OM-related tools have been created for this purpose. These tools enable LLM agents to simulate a traditional OM system, automating the entire matching process without human intervention. The overall infrastructure offers high scalability and allows extensive customisation. To the best of our knowledge, this study is the first to introduce an LLM-agent-based framework for OM tasks.

\section{Matching with LLM Agents}
\label{sec: approach}

Given a source ontology ($O_s$) and a target ontology ($O_t$), OM aims to find an alignment (A) that contains a set of pair-matched entities $\{(e_i, e_j)|e_i \in O_s, e_j \in O_t\}$. A classical matching process has two main steps: retrieval and matching. The retrieval step involves retrieving internal information from the ontology itself ($R_{int}$) and external information from a domain-specific thesaurus ($R_{ext}$). The matching step involves selecting the matching candidates ($M_{sel}$), running the matching algorithms ($M_{alg}$), and refining the matching results ($M_{ref}$). A classical OM can be formulated as:

\begin{equation}
R_{int} \Rightarrow R_{ext} \Rightarrow M_{sel} \Rightarrow M_{alg} \Rightarrow M_{ref}
\end{equation}

\begin{figure}[htbp]
\centering
\includegraphics[width=1\linewidth]{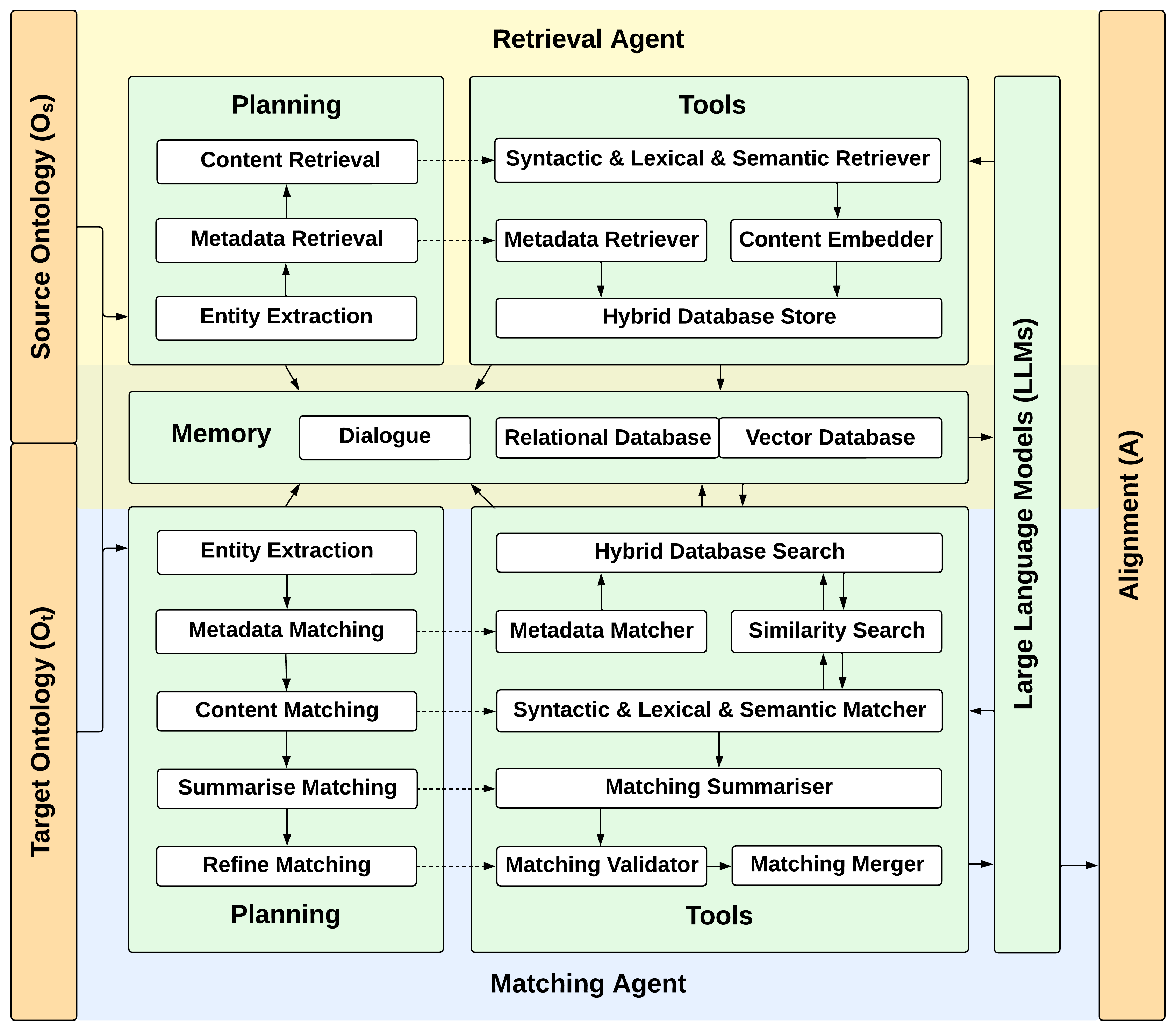}
\caption{Architecture of Agent-OM. All components are executed twice, once for each of the source and target ontologies, apart from the matching merger tool that combines the results from each pass.}
\label{fig: architecture}
\end{figure}

Figure~\ref{fig: architecture} shows the architecture of \emph{Agent-OM}, our agent-powered LLM-based OM framework. It retains the original input and output of the classical OM but modularises the two main steps with autonomous LLM agents, namely Retrieval Agent ($Agent\_{R}$) and Matching Agent ($Agent\_{M}$). We call these two LLM agents ``Siamese'' because they have their own planning modules and related tools but share memory. The memory is responsible for storing the information retrieved from the Retrieval Agent ($R_{sto}$) and facilitating the search of the stored information by the Matching Agent ($M_{sea}$). Therefore, an agent-based OM is formulated as:

\begin{equation}
Agent\_R_{(R_{int} \Rightarrow R_{ext} \Rightarrow R_{sto})} \Rightarrow Agent\_M_{(M_{sea} \Rightarrow M_{sel} \Rightarrow M_{alg} \Rightarrow M_{ref})}
\end{equation}

For each autonomous agent, the workflow is described as follows. The planning module decomposes a complex task into several subtasks and defines the order of subtasks and tools to be invoked. The plan is stored in the dialogue and passed to the LLMs. LLMs then invoke the tools to perform the subtasks. The tools may communicate with each other, with intermediate results stored in the dialogue. The tools can also access the database via the CRUD (create, read, update, and delete) functions provided. The entire workflow is driven by LLM prompts. We use solid lines to show the actual workflow controlled by the LLMs, and dotted lines to show the implicit link between a subtask and its corresponding tool activated by the LLMs.

\subsection{Retrieval Agent}

The Retrieval Agent is responsible for extracting entities from the ontologies, eliciting their metadata and ontology context information, and storing them in the hybrid database. For each entity extracted from the source and target ontologies, the planning module generates the instruction for retrieving the relevant information and feeding it into the LLMs to invoke the corresponding retrieval tools. The tools used in the Retrieval Agent include Metadata Retriever, Syntactic \& Lexical \& Semantic Retriever, and Hybrid Database Store with Content Embedder, described as follows.

\begin{itemize}[wide, noitemsep, topsep=0pt, labelindent=0pt]

\item Metadata Retriever ($R_{int}$): The metadata retriever collects the metadata of the input entity from the ontology, including its \emph{category} (i.e. either from the \emph{source} ontology or from the \emph{target} ontology) and \emph{type} (i.e. \emph{class} or \emph{property}).

\item Syntactic Retriever ($R_{int}$): The syntactic retriever is responsible for providing a unified text preprocessing result. A common text preprocessing pipeline consists of tokenisation, normalisation, stop words removal, and stemming/lemmatisation~\cite{manning2008introduction}. According to our previous study in~\cite{qiang2024does}, only tokenisation and normalisation help both matching completeness and correctness. The other two pipeline methods, stop words removal and stemming/lemmatisation, could cause unwanted false mappings. For this reason, our syntactic retriever considers only tokenisation and normalisation. We select white spaces to separate the words so that the outputs are short sentences that are easier for LLMs to interpret.

\item Lexical Retriever ($R_{int}$ \& $R_{ext}$): We consider three key aspects of the entity's lexical information: the general meaning ($R_{ext}$), the context meaning ($R_{ext}$), and the content meaning ($R_{int}$). In OM tasks, the general meaning is traditionally generated from Wikidata~\cite{vrandevcic2014wikidata} or similar corpus-based knowledge bases (KBs). As LLMs are trained from these KBs, we use the prompt ``What is the meaning of \{entity\_name\}?'' for the same function. However, using only the general meaning is not sufficient. Using the context constraint ``Context: \{context\}'' is effective in domain-specific tasks. Popular GPT-based domain applications, such as LawGPT~\cite{zhou2024lawgpt} and MedicalGPT~\cite{wang2023huatuo}, use similar approaches. Additionally, we also retrieve content information from \texttt{rdfs:label}, \texttt{rdfs:comment}, and other annotation properties, where the ontology creators may add comments or explanations. These are also useful for retrieving the meaning of the entity.

\item Semantic Retriever ($R_{int}$): The entity's semantic information includes its basic triple-based relations and more complex logic-based axioms. In this study, we only consider triple-based relations that can be verbalised into a more natural language-like presentation via a prompt-based verbalisation tool. Such verbalisation tools are not capable of handling complex logic-based axioms. These functions can only be achieved with external packages, such as OWL Verbaliser~\cite{owlverbaliser}, Sydney OWL Syntax~\cite{cregan2007sydney}, and the DeepOnto~\cite{he2023deeponto} verbalisation module~\cite{he2023language}.

\item Hybrid Database Store with Content Embedder ($R_{sto}$): We use a hybrid database system consisting of a traditional relational database and an advanced vector database. Entity metadata, such as the entity's category and type, are stored in the traditional relational database. In contrast, natural language-based content information, such as the entity's syntactic, lexical, and semantic information, is vectorised via a text embedding model and then stored in the vector database to enable similarity search based on relative distance in the vector space. A unique key links these two databases.

\end{itemize}

\subsection{Matching Agent}

The Matching Agent is responsible for finding possible correspondences, ranking and refining the results according to different criteria, and selecting the best matching candidate. For each entity extracted from the ontologies, the planning module generates the instruction for the matching types to be considered and feeds it into the LLMs to invoke the corresponding matching tools. The planning module first selects the source ontology as a starting point, extracting the entities from the ontology. Then, different matchers perform syntactic, lexical, or semantic matching functions to find the best match for the input entity, using a hybrid database search across the relational and vector databases. A predicted mapping is based on a summarised profile measure of syntactic matching, lexical matching, and semantic matching, with matching validation. The same procedure applies to the target ontology as a starting point, and the results of the common matching candidates are combined. The tools used in the Matching Agent include Hybrid Database Search, Metadata Matcher, Syntactic \& Lexical \& Semantic Matcher, Matching Summariser, Matching Validator, and Matching Merger, described as follows.

\begin{itemize}[wide, labelindent=0pt]

\item Hybrid Database Search ($M_{sea}$): The hybrid database search serves as an interface for the database accessible by the Metadata Matcher and Syntactic \& Lexical \& Semantic Matcher.

\item Metadata Matcher ($M_{sel}$): Given an input entity, the metadata matcher collects the category and type of the input entity from the relational database.

\item Syntactic \& Lexical \& Semantic Matcher ($M_{sel}$): Given an input entity, the syntactic \& lexical \& semantic matchers search for similar syntactic/lexical/semantic information respectively in the vector database using cosine similarity, defined for entities $\mathbf{A}$  and $\mathbf{B}$ as:

\begin{equation}
S_C(\mathbf{A}, \mathbf{B})
= \frac{\mathbf{A} \cdot \mathbf{B}}{\|\mathbf{A}\| \|\mathbf{B}\|}
= \frac{\sum_{i=1}^{n} \mathbf{A}_i \mathbf{B}_i}{\sqrt{\sum_{i=1}^{n} \mathbf{A}_i^2} \cdot \sqrt{\sum_{i=1}^{n} \mathbf{B}_i^2}}
\end{equation}

An extended search in the relational database is then used to filter the results based on the entity's metadata.

\item Matching Summariser ($M_{alg}$): We use reciprocal rank fusion (RRF)~\cite{cormack2009reciprocal} to summarise the matching results. Viewing each of the syntactic, lexical, and semantic descriptions as a document, the purpose of reciprocal rank is to accumulate the inverse of the ranks $r$ of documents $d$ over three ranking results from syntactic, lexical, and semantic matching, defined as:

\begin{equation}
RRF(d \in D) = \sum_{r\in R} \frac{1}{k + r(d)}
\end{equation}

$k$ is a constant parameter that is conventionally set to 0 as we do here. This ensures that the formula most highly rewards the most highly-ranked entities. In our case, we are evaluating each entity that occurs in the top@k of each of the three rankings (i.e. syntactic matching, lexical matching, and semantic matching), and combining their results as an overall matching summary.

\item Matching Validator ($M_{ref}$): Validation is a critical step in minimising LLM hallucinations, as illustrated in SelfCheckGPT~\cite{manakul2023selfcheckgpt}. We also apply this method to the summarised results. We ask the LLM a binary question ``Question: Is \{entity\_name\} equivalent to \{matching\_entity\_name\}? Context: \{context\} Answer the question within the context. Answer yes or no. Give a short explanation.'' to check whether the predicted entity is equivalent to the input entity in the provided context. For computational efficiency, we iterate the comparison from rank 1 to $n$ and select the highest-ranked match with a "yes" answer for the matching merger.

\item Matching Merger ($M_{ref}$): The matching merger is responsible for combining the results from a search of the source ontology and a search of the target ontology. In this study, we select only the correspondences found on both sides. As an agent-based system, this can be extended to use multi-agent negotiation via the correspondence inclusion dialogue~\cite{payne2014negotiating}.

\end{itemize}

\section{Implementation Details}
\label{sec: implementation}

We implement our design of the framework in a proof-of-concept system. The components and their implementation are as follows:

\begin{itemize}[wide, noitemsep, topsep=0pt, labelindent=0pt]

\item LLMs: Our system supports a wide range of LLMs, including OpenAI GPT~\cite{llm-gpt}, Anthropic Claude~\cite{llm-claude}, Meta Llama~\cite{llm-llama}, Alibaba Qwen~\cite{llm-qwen}, Google Gemma~\cite{llm-gemma}, and ChatGLM~\cite{glm2024chatglm}. We select 10 models for this study. 4 models are API-accessed commercial LLMs, while the other 6 are open-source LLMs. Table~\ref{tab: llm-version} gives the details. For API-accessed LLMs, we include two models of different sizes for each family of models. For open-source LLMs, we select models with similar sizes (7-9 billion parameters) from different families. They are accessed via the Ollama library~\cite{ollama}.

\begin{table}[htbp]
\centering
\renewcommand\arraystretch{1.2}
\tabcolsep=0.15cm
\caption{Details of LLMs used in the study.}
\label{tab: llm-version}
\begin{adjustbox}{width=1\columnwidth,center}
\begin{tabular}{|c|l|c|l|} 
\hline
\multicolumn{1}{|c|}{\textbf{Family}} & \multicolumn{1}{|l|}{\textbf{Model}}& \multicolumn{1}{|c|}{\textbf{Size}}   & \multicolumn{1}{|l|}{\textbf{Version}}    \\ \hline 
\multirow{2}{*}{GPT}                  & gpt-4o                              & N/A       & gpt-4o-2024-05-13                 \\ \cline{2-4}
                                      & gpt-4o-mini                         & N/A       & gpt-4o-mini-2024-07-18            \\ \cline{1-4}
\multirow{2}{*}{Claude}               & claude-3-sonnet                     & N/A       & claude-3-sonnet-20240229          \\ \cline{2-4}
                                      & claude-3-haiku                      & N/A       & claude-3-haiku-20240307           \\ \cline{1-4}
\multirow{2}{*}{Llama}                & llama-3-8b*                         & 4.7 GB    & Ollama Model ID: 365c0bd3c000     \\ \cline{2-4}
                                      & llama-3.1-8b*                       & 4.9 GB    & Ollama Model ID: 46e0c10c039e     \\ \cline{1-4}
\multirow{2}{*}{Qwen}                 & qwen-2-7b*                          & 4.4 GB    & Ollama Model ID: dd314f039b9d     \\ \cline{2-4}
                                      & qwen-2.5-7b*                        & 4.7 GB    & Ollama Model ID: 845dbda0ea48     \\ \cline{1-4}       
\multirow{1}{*}{Gemma}                & gemma-2-9b*                         & 5.4 GB    & Ollama Model ID: ff02c3702f32     \\ \cline{1-4}
\multirow{1}{*}{GLM}                  & glm-4-9b*                           & 5.5 GB    & Ollama Model ID: 5b699761eca5     \\ \hline
\multicolumn{4}{c}{* Open-source LLM (retrieved December 1, 2024).} \\
\end{tabular}
\end{adjustbox}
\end{table}

\item Planning: We select the LangChain library~\cite{langchain}. The library provides a wide range of agents. We select the tool calling agent (a.k.a. function calling agent). At the time of writing, the LangChain library only supports this type of agent used with commercial API-accessed LLMs. To extend our framework to open-source LLMs, we employ the similar concept of ``chain'' to simulate the tool calling agent for open-source LLMs.

\item Memory: (1) Short-term memory: We use a conversational dialogue to store the original intermediate output of each operating process, with no map-reduce applied. (2) Long-term memory: We select a hybrid database consisting of a traditional relational database and an advanced vector database. PostgreSQL~\cite{postgresql} supports a standalone integration of the traditional relational database and the extended vector database using pgvector~\cite{pgvector}. We select OpenAI embedding models~\cite{openaiembeddingmodels} for the content embedding in the vector database. Alternatives are Google Vertex AI Embeddings~\cite{vertexaiembeddings} or Sentence-BERT~\cite{reimers2019sentencebert}, but the dimension of the embedding changes between different embedding models.

\item Tools: To demonstrate the flexibility of our framework, we present the usage of prompt-based tools and programming-based tools, as well as the tools that combine a mixture of prompt-based and programming-based tools.

\end{itemize}

\subsection{Ontology Naming Conventions}
\label{sec: naming conventions}

In this work, the term \textit{entity} is a general expression for ontology classes or properties (without specifying which). We use \textit{entity uri} to mean a fully expanded class name or property name with respect to its prefix. We use \textit{entity name} to mean a class name or property name without its prefix. For example, the \textit{entity uri} is ``http://cmt\#ProgramCommitteeChair'' and the \textit{entity name} is ``ProgramCommitteeChair''.

Naming conventions for entities fall into two types: the name has a natural language meaning (Type 1); or the name is a code (Type 2). We observe that LLMs can perform well with meaningful entity names (e.g. ProgramCommitteeChair and Chair\_PC). Often in larger biomedical ontologies, entities are codes and their meaningful descriptions are in their labels or comments (e.g. MA\_0000270 and NCI\_C33736). For this type of naming convention, current LLMs tend to generate the wrong synthesised label or comment corresponding to the code. For example, LLMs can mistakenly interpret the codes ``MA\_0000270'' and ``NCI\_C33736'' both to be ``liver'', while the intended meanings of these two codes are ``eyelid tarsus'' and ``Tarsal\_Plate''.

To handle the variety of ontology naming conventions and standardise their usage in LLM-based OM, we use a unified naming convention in this study. If the entity is a code, we use its label or comment instead. For example, we use ``eyelid tarsus'' and ``Tarsal\_Plate'' instead of ``MA\_0000270'' and ``NCI\_C33736'', respectively. In case the two ontologies reuse the same entity name, we assign a prefix to each entity with its serial number, category, and type. For example, if it were the case that ``ProgramCommitteeChair'' appears in both source and target ontologies, the unique identifier for each entity would be ``023-Source-Class-ProgramCommitteeChair'' and ``042-Target-Class-ProgramCommitteeChair''.

\subsection{Running Example}
\label{sec: a running example}

To demonstrate the usability of our framework, we choose the CMT-ConfOf alignment as a sample alignment. The CMT Ontology is the source ontology and the ConfOf Ontology is the target ontology. Both ontologies contain similar concepts related to conference organisation. The running example aims to find the best matching entity in the target ontology corresponding to the entity ``http://cmt\#ProgramCommitteeChair'' in the source ontology.

\subsubsection{Retrieval Agent}

Table~\ref{tab: retriever} illustrates the tool calling in the Retrieval Agent. For the entity ``http://cmt\#ProgramCommitteeChair'', the agent first calls the Metadata Retriever to find its category and type. Then the Syntactic Retriever is invoked, giving the output ``program committee chair''. The agent next invokes the Lexical Retriever to generate a detailed description: ``In the context of a conference, `ProgramCommitteeChair' refers to...''. The Semantic Retriever generates related triple relations, such as ``ProgramCommitteeChair rdfs:subClassOf ProgramCommitteeMember''. These triples are verbalised using natural language: ``The class `ProgramCommitteeChair' is a subclass of `ProgramCommitteeMember'.''

\begin{figure}[htbp]
\centering
\includegraphics[width=1\linewidth]{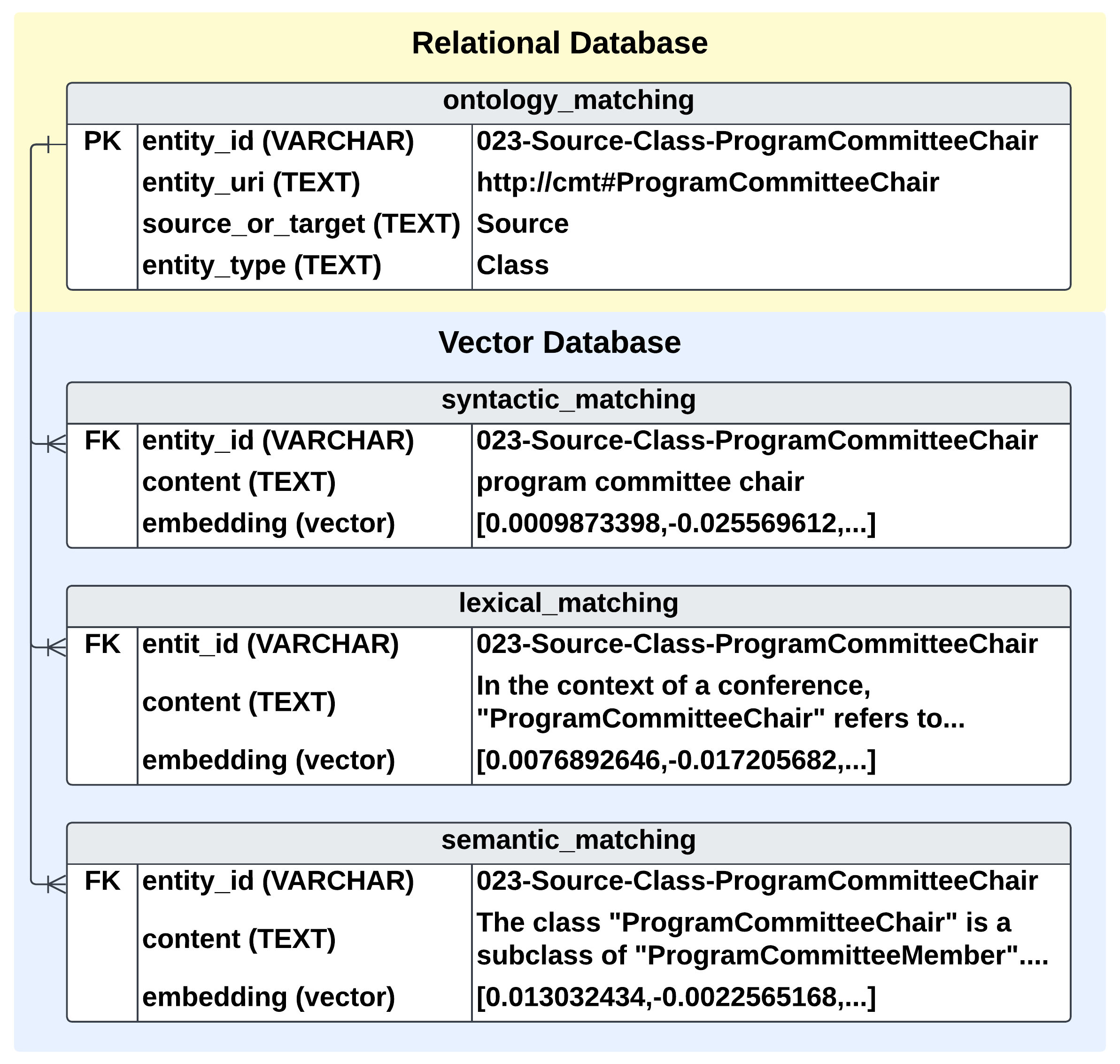}
\caption{Storing \textit{``http://cmt\#ProgramCommitteeChair''}.}
\label{fig: database}
\end{figure}

While each entity has its own syntactic, lexical, and semantic information, a naive approach to deciding if two entities are the same is to generate a binary question for every pair of entities as a prompt to the LLM: ``Is Entity1 equivalent to Entity2? Consider the following: The syntactic information of Entity1 is... The lexical information of Entity1 is... The semantic information of Entity1 is... The syntactic information of Entity2 is... The lexical information of Entity2 is... The semantic information of Entity2 is... '' This approach has two limitations. (1) LLMs have token limits that restrict the number of tokens processed for each interaction. Combining all the retrieved information may exceed token limits. (2) The binary comparison is costly because the complexity of the comparison is the product of the number of entities in the source ontology and the target ontology. We bypass these limitations by (1) using an open question instead and (2) storing useful information in a searchable database. Figure~\ref{fig: database} shows the entity metadata and content information stored in the relational database and the vector database, respectively. On one hand, the entity's metadata is needed to find an exact match. For example, ``http://cmt\#ProgramCommitteeChair'' is a class in the source ontology, so the matched entity should be a class in the target ontology. On the other hand, content information including an entity's syntactic, lexical, and semantic information is used for a similarity-based match because they are usually retrieved as natural language, which can be more ambiguous than metadata. Similarity between natural language terms is commonly based on embedding vectors,  for which the vector database enables fast similarity searches.

\begin{table*}[htbp]
\centering
\renewcommand\arraystretch{1.2}
\tabcolsep=0.15cm
\caption{Tool calling in the Retrieval Agent.}
\label{tab: retriever}
\begin{adjustbox}{height=0.88\height,center}
\begin{tabular}{p{1.12\linewidth}}
\hline
\textbf{Tool: Metadata Retriever}    \\ \hline
\textbf{Input:} entity\_uri = ``http://cmt\#ProgramCommitteeChair'' \\
\textbf{Extract:}  source\_or\_target = ``Source'', entity\_type = ``Class'' \\ \hline
\hline
\textbf{Tool: Syntactic Retriever}    \\ \hline
\textbf{Input:} entity\_uri = ``http://cmt\#ProgramCommitteeChair''  \\
\textbf{Extract:} entity\_name = ``ProgramCommitteeChair'' from entity\_uri.  \\
\textbf{Method:} tokenise\_and\_normalise(entity\_name)  \\
\textbf{Output:} entity\_syntactic = ``program committee chair'' \\ \hline
\hline
\textbf{Tool: Lexical Retriever}    \\ \hline
\textbf{Input:} entity\_uri = ``http://cmt\#ProgramCommitteeChair'', context = ``conference''  \\
\textbf{Extract:} entity\_name = ``ProgramCommitteeChair'' from entity\_uri.  \\
extra\_information from annotation properties related to entity\_uri. \\
\textbf{Prompt:} Question: What is the meaning of \{entity\_name\}? Context: \{context\} Extra Information: \{extra\_information\}  \\
Answer the question within the context and using the extra information.  \\
\textbf{Output:} entity\_lexical = ``In the context of a conference, `ProgramCommitteeChair' refers to...'' (AI-generated content)  \\ \hline
\hline
\textbf{Tool: Semantic Retriever}    \\ \hline
\textbf{Input:} entity\_uri = ``http://cmt\#ProgramCommitteeChair'' \\
\textbf{Method:} generate\_subgraph(entity\_uri) \\
\textbf{Output:} entity\_subgraph consists of triples related to entity\_uri. \\
\textbf{Prompt:} Verbalise triples into phrases: \{entity\_subgraph\} \\
\textbf{Output:} entity\_semantic = ``The class `ProgramCommitteeChair' is a subclass of `ProgramCommitteeMember'...'' (AI-generated content)  \\ \hline
\hline
\textbf{Tool: Hybrid Database Store with Content Embedder}    \\ \hline
\textbf{Input:} entity\_uri = ``http://cmt\#ProgramCommitteeChair'', source\_or\_target = ``Source'', entity\_type = ``Class'' \\
\textbf{Extract:} entity\_id = ``023-Source-Class-ProgramCommitteeChair''  \\
\textbf{Query: Create a relational database and store entity's metadata} 
\begin{lstlisting}[language=SQL]
DROP TABLE IF EXISTS ontology_matching CASCADE;
CREATE TABLE ontology_matching (entity_id VARCHAR(1024) PRIMARY KEY, entity_uri TEXT, source_or_target TEXT, entity_type TEXT);
INSERT INTO ontology_matching (entity_id, entity_uri, source_or_target, entity_type)
VALUES ({entity_id}, {entity_uri}, {source_or_target}, {entity_type});
\end{lstlisting}
\textbf{Input}: entity\_syntactic = ``program committee chair'', entity\_lexical = ``In the context of a conference, `ProgramCommitteeChair' refers to...'',    \\
entity\_semantic = ``The class `ProgramCommitteeChair' is a subclass of `ProgramCommitteeMember'...'',  \\
matching\_table = "syntactic\_matching/lexical\_matching/semantic\_matching"   \\
\textbf{Extract}: content\_embedding based on entity\_syntactic/entity\_lexical/entity\_semantic.  \\
\textbf{Query: Create a vector database and store entity's syntactic, lexical, and semantic information}
\begin{lstlisting}[language=SQL]
CREATE EXTENSION IF NOT EXISTS vector;
DROP TABLE IF EXISTS {matching_table};
CREATE TABLE {matching_table} 
(entity_id VARCHAR(1024) NOT NULL REFERENCES ontology_matching(entity_id), content TEXT, embedding vector(1536));
INSERT INTO {matching_table} (entity_id, content, embedding)
VALUES ({entity_id}, {entity_syntactic}/{entity_lexical}/{entity_semantic}, {content_embedding});
\end{lstlisting}
\textbf{Output:} One relational database table (ontology\_matching) and three vector database tables (syntactic\_matching, lexical\_matching, and semantic\_matching). \\ \hline
\end{tabular}
\end{adjustbox}
\end{table*}

\subsubsection{Matching Agent}

Table~\ref{tab: matcher} demonstrates the tool calling in the Matching Agent. Given ``http://cmt\#ProgramCommitteeChair'' from the source ontology, the matching entities found by each matcher are stored in the dialogue and combined using the RRF function. The result of the Matching Summariser is a list of predicted mappings. The next step is to refine the predicted mappings. The Matching Validator asks a binary question to compare whether the given entity is the same or different to the predicted matching entity in RRF-descending order. Because the validator receives a ``yes" answer for the first iteration of the entity ``http://confOf\#Chair\_PC'', the Matching Agent outputs ``http://confOf\#Chair\_PC'' as the best matching entity found in the target ontology. The Matching Merger combines the results from the same procedure applied in the search from ``http://confOf\#Chair\_PC'' in the target ontology. These two terms are considered as matched entities only if the mapping can be found bidirectionally (i.e. ``http://cmt\#ProgramCommitteeChair'' is also found to be the best matching entity in the source ontology for the search from ``http://confOf\#Chair\_PC'' in the target ontology).

\begin{table*}[htbp]
\centering
\renewcommand\arraystretch{1.2}
\tabcolsep=0.15cm
\caption{Tool calling in the Matching Agent.}
\label{tab: matcher}
\begin{adjustbox}{height=0.88\height,center}
\begin{tabular}{p{1.12\linewidth}}
\hline
\textbf{Tool: Hybrid Database Search with Metadata Matcher}    \\ \hline
\textbf{Input:} entity\_uri = ``http://cmt\#ProgramCommitteeChair'', source\_or\_target = ``Source''\\
matching\_table = ``syntactic\_matching/lexical\_matching/semantic\_matching'' \\
\textbf{Query: Find entity id} 
\begin{lstlisting}[language=SQL]
SELECT o.entity_id FROM ontology_matching o
WHERE o.entity_uri = {entity_uri} and o.source_or_target = {source_or_target};
\end{lstlisting}
\textbf{Output:} entity\_id = ``023-Source-Class-ProgramCommitteeChair''\\
\textbf{Query: Find entity metadata} 
\begin{lstlisting}[language=SQL]
SELECT o.entity_type, m.embedding 
From ontology_matching o, {matching_table} m
WHERE o.entity_id = m.entity_id AND o.entity_id = {entity_id};
\end{lstlisting}
\textbf{Output:} entity\_type = ``Class'', content\_embedding = [...]  \\ \hline
\hline
\textbf{Tool: Hybrid Database Search with Syntactic \& Lexical \& Semantic Matcher}    \\ \hline
\textbf{Input:} content\_embedding = [...], matching\_table = ``syntactic\_matching/lexical\_matching/semantic\_matching'', \\
similarity\_threshold = 0.90, top\_k = 3, source\_or\_target = ``Source'', entity\_type = ``Class'' \\
\textbf{Query: Find matching candidates} 
\begin{lstlisting}[language=SQL]
WITH vector_matches AS (
SELECT entity_id, 1 - (embedding <=> '{content_embedding}') AS similarity
FROM {matching_table}
WHERE 1 - (embedding <=> '{content_embedding}') >= {similarity_threshold})
SELECT o.entity_id, v.similarity as similarity 
FROM ontology_matching o, vector_matches v
WHERE o.entity_id IN (SELECT entity_id FROM vector_matches) 
AND o.entity_id =  v.entity_id AND o.source_or_target != {source_or_target} AND o.entity_type = {entity_type}
ORDER BY v.similarity DESC
LIMIT {top_k};
\end{lstlisting}
\textbf{Output:} matching\_results = [(``syntactic\_matching'', []), (``lexical\_matching'', [``095-Target-Class-Chair\_PC'']), 
(``semantic\_matching'', [``103-Target-Class-Member\_PC'', ``114-Target-Class-Scholar'', ``092-Target-Class-Author''])]\\ \hline
\hline
\textbf{Tool: Matching Summariser}    \\ \hline
\textbf{Input:} matching\_results = [...] \\
\textbf{Method:} reciprocal\_rank\_fusion(matching\_results)\\
\textbf{Output:} matching\_summary = [(1.0, [``095-Target-Class-Chair\_PC'', ``103-Target-Class-Member\_PC'']), (0.5, [``114-Target-Class-Scholar'']), (0.33, [``092-Target-Class-Author''])] \\ \hline
\hline
\textbf{Tool: Matching Validator}    \\ \hline
\textbf{Input:} entity\_uri = ``http://cmt\#ProgramCommitteeChair'', matching\_summary = [...], context = ``conference'' \\
\textbf{Extract:} entity\_name from entity\_uri, each matching\_entity\_name from matching\_summary.\\
\textbf{Prompt:} Question: Is \{entity\_name\} equivalent to \{matching\_entity\_name\}? Context: \{context\} \\
Answer the question within the context. Answer yes or no. Give a short explanation. \\
\textbf{Output:} best\_matching\_entity\_id = ``095-Target-Class-Chair\_PC''\\
``095-Target-Class-Chair\_PC'': ``Yes. In the context of a conference, the term `program committee chair' is equivalent to `chair PC.' Both refer to the individual responsible for overseeing the program committee, which is tasked with organizing the conference's academic or technical program, including the review and selection of submitted papers.'' (AI-generated content) \\
``103-Target-Class-Member\_PC'': ``No. The program committee chair is not equivalent to a member of the program committee (PC). The program committee chair is responsible for overseeing the entire review process, coordinating the activities of the PC members, and making final decisions on the acceptance of submissions. In contrast, a member of the PC is primarily responsible for reviewing and evaluating submitted papers.'' (AI-generated content) \\
\textbf{Query: Find best matching entity} 
\begin{lstlisting}[language=SQL]
SELECT o.entity_uri FROM ontology_matching o
WHERE o.entity_id = {best_matching_entity_id};
\end{lstlisting}
\textbf{Output:} best\_matching\_entity\_uri = ``http://confOf\#Chair\_PC''\\ \hline
\hline
\textbf{Tool: Matching Merger}    \\ \hline
\textbf{Output:} Merge the best matching entity in the source ontology for the search from ``http://confOf\#Chair\_PC'' in the target ontology. \\ \hline
\end{tabular}
\end{adjustbox}
\end{table*}

\section{Evaluation}
\label{sec: evaluation}

\subsection{Evaluation Criteria}

In information retrieval, a common assessment of matching tasks is based on comparing predicted results with the expected output. Precision and recall are used to measure the correctness and completeness of the matching, respectively. When adapting these measures to OM, the predicted results generated by the system are denoted Alignment (A), and the expected results provided by the domain experts are denoted Reference (R)~\cite{do2003comparison}. Therefore, precision and recall for OM tasks are defined as:

\begin{equation}
Precision = \frac{|A \cap R|}{|A|} \qquad Recall = \frac{|A \cap R|}{|R|}
\end{equation}

Precision and recall are commonly combined into a single measure F1 score, defined as:

\begin{equation}
F_1 \ Score= \frac{2}{Precision^{-1} + Recall^{-1}}
\end{equation}

\subsection{Evaluation of Three OAEI Tracks}
\label{sub-sec: OAEI evaluation}

In this section, we test our proof-of-concept system with three OAEI tracks containing different types of OM tasks. These include few-shot tasks with a small proportion of trivial correspondences (\ref{evaluation: conference-track} Test Case), simple tasks with a large proportion of trivial correspondences (\ref{evaluation: anatomy-track} Test Case 1 and~\ref{evaluation: mse-track} Test Case 3), complex tasks with a large proportion of non-trivial correspondences (\ref{evaluation: anatomy-track} Test Case 2), with complex references (\ref{evaluation: mse-track} Test Case 1), or requiring domain-specific knowledge (\ref{evaluation: mse-track} Test Case 2). We report the evaluation metrics for the best-performing singular model gpt-4o and its hyperparameter settings over a single run. We ran multiple trials and found slight differences in the results due to the non-determinism of LLMs, but these differences are not significant with respect to the precision of the results we report. For all test cases in the three OAEI tracks, we select the hyperparameter settings of similarity\_threshold = 0.90 and top@k = 3. See Section~\ref{ablation: hyperparameter} for a discussion on the hyperparameter settings of Agent-OM.

\subsubsection{OAEI Conference Track}
\label{evaluation: conference-track}

The OAEI Conference Track contains pairwise alignments for 7 small and medium-sized conference-related ontologies with a total of 21 matching tasks~\cite{cheatham2014conference,solimando2014detecting,solimando2017minimizing,zamazal2017ten}. In each alignment, the trial correspondences that can be used to train the models are very limited (commonly less than 10). All conference ontologies in this track use the Type 1 naming convention, where the names of classes and properties have meanings. In this study, we use the publicly available reference ra1-M3 as the reference (R), including class and property mappings.

\begin{figure}[htbp]
\centering
\includegraphics[width=1\linewidth]{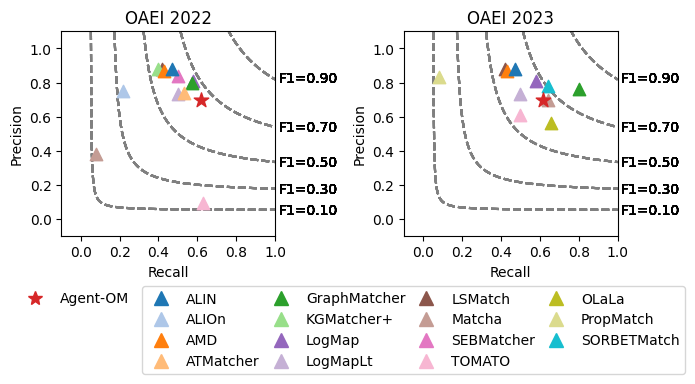}
\caption{OAEI Conference Track Test Case.}
\label{fig: conference-domain}
\end{figure}

Figure~\ref{fig: conference-domain} compares Agent-OM with the 15 OM systems in OAEI 2022 and OAEI 2023. Agent-OM achieves above-average performance. Its overall F1 score ranks 3/13 in 2022 and 5/12 in 2023. We note that ra1-M3 is known to be missing valid equivalence mappings. We believe that Agent-OM could achieve better performance over a complete reference such as ra2-M3 or rar2-M3. These are not publicly available at the time of writing.

\subsubsection{OAEI Anatomy Track}
\label{evaluation: anatomy-track}

The OAEI Anatomy Track contains a reference alignment for the mouse anatomy and the human anatomy, created and evolved from~\cite{bodenreider2005mice,beisswanger2012towards,euzenat2011ontology,dragisic2017experiences}. Both ontologies use the Type 2 naming convention, where the names of classes and properties are biomedical codes. We report the results of our evaluation in two parts: alignment with trivial correspondences and alignment with non-trivial correspondences.

\begin{enumerate}[wide, noitemsep, topsep=0pt, labelindent=0pt]

\item Test Case 1: The track originally contains a large proportion of trivial correspondences that have the same standardised labels (e.g. ``femoral artery'' and ``Femoral\_Artery''). Figure~\ref{fig: anatomy-easy} compares Agent-OM with the results of the 12 OM systems in OAEI 2022 and OAEI 2023 for alignment with trivial correspondences. Almost all OM systems achieve relatively high precision and recall when matching with trivial correspondences. For Agent-OM, its overall F1 score ranks the second highest in 2022 and 2023.

\begin{figure}[htbp]
\centering
\includegraphics[width=1\linewidth]{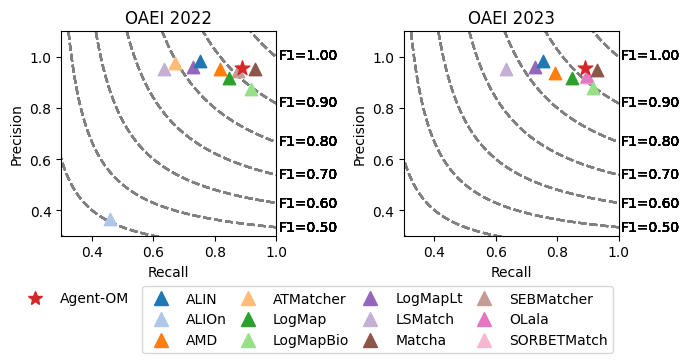}
\caption{OAEI Anatomy Track Test Case 1.}
\label{fig: anatomy-easy}
\end{figure}

\item Test Case 2: We remove these trivial correspondences from both the reference (R) and alignment (A) to focus the matching performance comparison on non-trivial correspondences. Figure~\ref{fig: anatomy-hard} compares Agent-OM with the results of the 12 OM systems in OAEI 2022 and OAEI 2023 for alignment with non-trivial correspondences. We observe better performance by Agent-OM. The overall F1 score of Agent-OM is superior to the 11 OM systems including an LLM-based OM system OLaLa, and only behind one deep learning (DL)-based OM system Matcha~\cite{cotovio2024matcha}, which could have benefited from an unusually large training set available for this case.

\begin{figure}[htbp]
\centering
\includegraphics[width=1\linewidth]{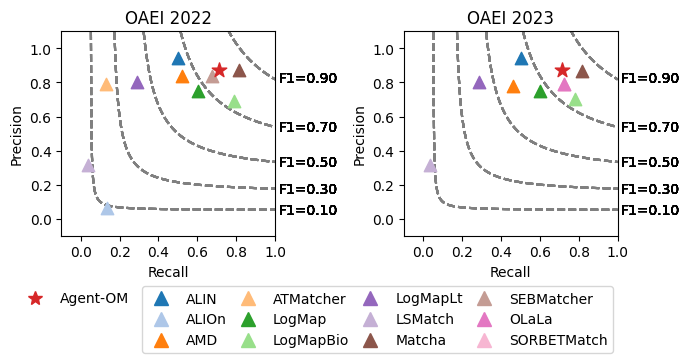}
\caption{OAEI Anatomy Track Test Case 2.}
\label{fig: anatomy-hard}
\end{figure}

\end{enumerate}

\subsubsection{OAEI MSE Track}
\label{evaluation: mse-track}

The OAEI MSE Track provides reference alignments for ontologies in materials science and engineering~\cite{engy2020}. The track contains three test cases aligning MaterialInformation~\cite{ashino2010materials}, MatOnto~\cite{matonto}, and EMMO~\cite{emmo}. The MaterialInformation and MatOnto use the Type 1 naming convention, while the EMMO uses the Type 2 naming convention.

\begin{enumerate}[wide, noitemsep, topsep=0pt, labelindent=0pt]

\item Test Case 1: This test case provides a reference alignment for a fragment of MaterialInformation and the medium-sized MatOnto. The challenge of this task arises due to the reference intentionally including several subsumption correspondences, when OM systems may mistakenly map subsumptions into equivalence relations. Figure~\ref{fig: mse-1} compares Agent-OM with the results of the 4 OM systems in OAEI 2022 and OAEI 2023. Agent-OM achieves the best performance, with the highest F1 score.

\begin{figure}[htbp]
\centering
\includegraphics[width=1\linewidth]{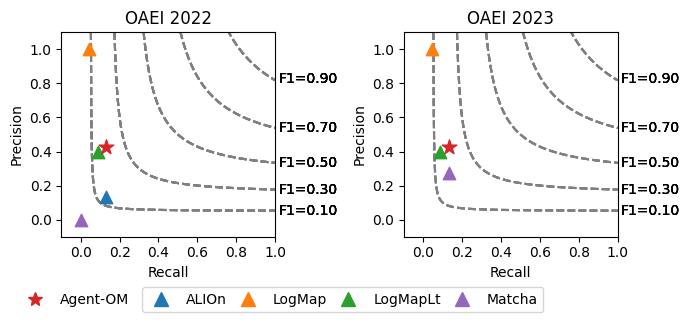}
\caption{OAEI MSE Track Test Case 1.}
\label{fig: mse-1}
\end{figure}

\item Test Case 2: This test case provides a reference alignment for the complete MaterialInformation and the medium-sized MatOnto. The challenge of this task is to align many examples of specific terminology, abbreviations, and acronyms used in materials science. For example, ``Au'' stands for ``Gold'', ``Ag'' stands for ``Silver'', and ``Cu'' stands for ``Copper''. Figure~\ref{fig: mse-2} compares Agent-OM with the results of the 4 OM systems in OAEI 2022 and OAEI 2023. Agent-OM achieves the best performance in precision, recall, and overall F1 score across the results of OAEI 2022 and OAEI 2023. We should expect an LLM-based matcher to have high recall on this test case due to its access to extensive training literature.

\begin{figure}[htbp]
\centering
\includegraphics[width=1\linewidth]{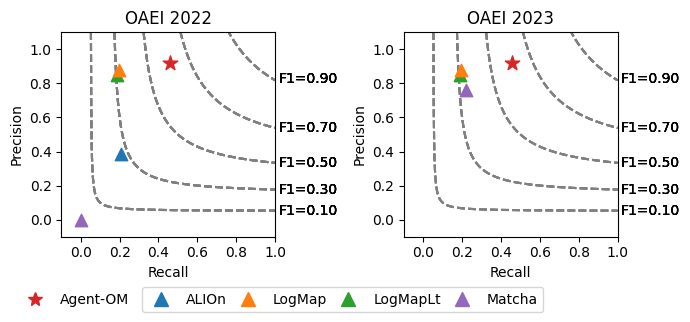}
\caption{OAEI MSE Track Test Case 2.}
\label{fig: mse-2}
\end{figure}

\item Test Case 3: This test case provides a reference alignment for the complete MaterialInformation and the medium-sized EMMO. The EMMO extends the upper ontology called Basic Formal Ontology (BFO). This means that the classes in the EMMO are somewhat standardised according to the BFO classes. Figure~\ref{fig: mse-3} compares Agent-OM with the results of the 4 OM systems in OAEI 2022 and OAEI 2023. The performance of Agent-OM is competitive with the best of the OAEI 2022 results and in OAEI 2023 is bettered by the DL-based OM system Matcha.

\begin{figure}[htbp]
\centering
\includegraphics[width=1\linewidth]{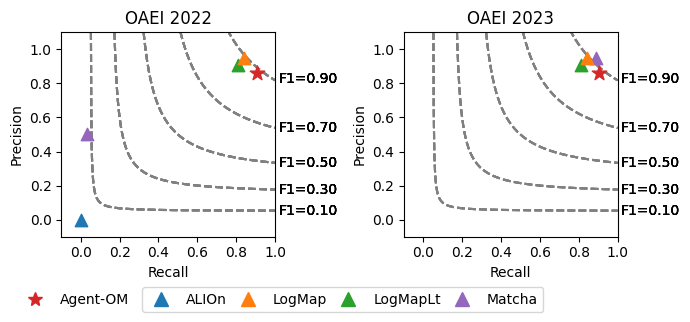}
\caption{OAEI MSE Track Test Case 3.}
\label{fig: mse-3}
\end{figure}

\end{enumerate}

\section{Ablation Study}
\label{sec: ablation}

\subsection{System Components}

\subsubsection{Architectures}
\label{ablation: architectures}

We compare Agent-OM with two simpler architectures where the OM is much more reliant on straightforward LLM use. (1) \textit{LLM-Only}: Given $O_s$ and $O_t$, this approach extracts each $e1 \in O_s$ and $e2 \in O_t$. The matching decision is purely based on LLMs without any additional information. (2) \textit{LLM-with-Context}: Given $O_s$ and $O_t$, this approach extracts each $e1 \in O_s, e2 \in O_t$ and their syntactic, lexical, and semantic information. The matching decision uses LLMs to determine whether two concepts are identical or not based on the information provided.

\begin{figure}[htbp]
\centering
\includegraphics[width=1\linewidth]{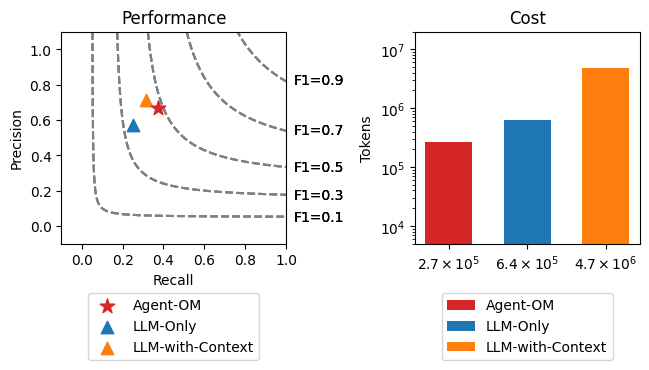}
\caption{Comparison with LLM-based architectures.}
\label{fig: different-architecture}
\end{figure}

The experiment is run on the CMT-ConfOf alignment demonstrated in Section~\ref{sec: a running example}. We use the GPT model gpt-4o. For Agent-OM, we choose the hyperparameter settings of similarity\_threshold = 0.90 and top@k = 3. Figure~\ref{fig: different-architecture} compares Agent-OM with \textit{LLM-Only} and \textit{LLM-with-Context}. \textit{LLM-Only} shows low precision and recall. \textit{LLM-with-Context} partially overcomes this deficiency by providing additional information, but token consumption is extremely high without optimising the matching candidate selection. The performance of \textit{LLM-with-Context} is unstable over multiple runs due to the effects of stochastic LLM results. Our Agent-OM architecture handles these challenges with tool calling agents and hybrid database searches. Note that the CMT-ConfOf alignment demonstrated here is a small alignment task, while Agent-OM is expected to be relatively more effective and efficient in large-scale OM tasks.

\subsubsection{LLMs}
\label{ablation: llm}

\begin{figure}[htbp]
\centering
\includegraphics[width=1\linewidth]{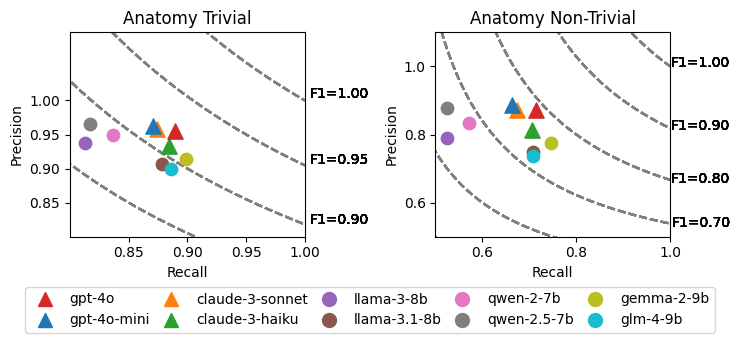}
\caption{Comparison of different LLMs. API-accessed models are shown as triangles and open-source models as circles.}
\label{fig: different-llm}
\end{figure}

Figure~\ref{fig: different-llm} varies LLMs in Agent-OM on the OAEI Anatomy Track. In general, API-accessed models perform better than open-source models. The leading models, gpt-4o and claude-3-sonnet, are both large API-accessed models. Among open-source models, gemma-2-9b achieves the best performance, while llama-3-8b is relatively poor. Curiously, although we see improved performance with llama-3.1-8b over its previous version, qwen-2.5-7b does not show an advantage over its previous version. This may be a side effect of LLM developers optimising for tasks other than OM. We experimented with other LLMs derived from the Llama and Qwen families and generally found poor performance, possibly due to the fine-tuning for specific tasks.

\subsubsection{Text Embedding Models}

We also test three different text embeddings in~\cite{openaiembeddingmodels} on the OAEI Anatomy Track. The default length of the embedding vector is 1536 for text-embedding-ada-002 and text-embedding-3-small, and 3072 for text-embedding-3-large. We do not observe a significant difference arising from varying the text embeddings from text-embedding-3-small to text-embedding-3-large. We do not see that text-embedding-3-small and text-embedding-3-large perform better than text-embedding-ada-002.

\subsubsection{Hybrid Database} 

The use of a hybrid database unlocks the potential for search-based OM. We define $N_s$ and $N_t$ as the number of entities extracted from the source ontology ($O_s$) and the target ontology ($O_t$), respectively. In naive LLM-based OM, the matching is performed using binary questions to compare each pair of entities from $O_s$ and $O_t$ based on their relevant information. The complexity of the comparison is $N_s \times N_t$ (for retrieval and matching). In search-based OM, we first retrieve entity information from $O_s$ and $O_t$ and store it in a hybrid database. Following our design and implementation in Section~\ref{sec: approach} and~\ref{sec: implementation}, the complexity is $N_s+N_t$ (for retrieval) + 0 (for search) + $k(N_s+N_t)$ (for validation) + 0 (for merge). Search-based OM is cost-effective because the inequality $(k+1)(N_s+N_t)<N_s \times N_t$ always holds in common OM task settings where $N_s, N_t \gg k+1$. We also apply two tools to reduce LLM hallucinations: the matching validator and the matching merger.

\subsubsection{Matching Validator} 

\begin{figure}[htbp]
\centering
\subfloat{\includegraphics[width=0.333\columnwidth]{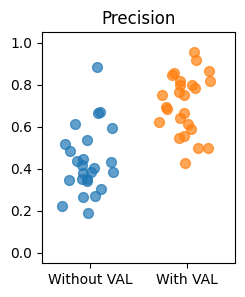}}
\subfloat{\includegraphics[width=0.333\columnwidth]{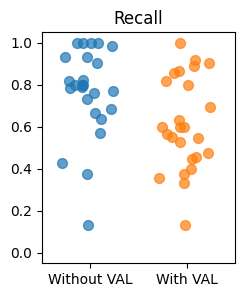}}
\subfloat{\includegraphics[width=0.333\columnwidth]{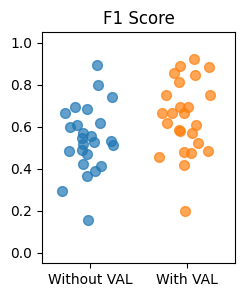}}
\caption{Comparison of without/with matching validator.}
\label{fig: validation}
\end{figure}

We employ a matching validator by asking the LLM to self-check the candidate correspondences. It is helpful in detecting two common types of false positive mappings: (1) non-existent mappings and (2) counter-intuitive mappings. Figure~\ref{fig: validation} compares precision, recall, and F1 score with and without validation in the three OAEI tracks we analysed. The matching results with validation generally achieved an improvement in precision and F1 score, with a slight decrease in recall. This is in line with the findings in CoT with self-consistency (CoT-SC)~\cite{wang2023selfconsistency}, where the provision of a self-check can reduce LLM hallucinations. Our experiment set the hyperparameters to similarity\_threshold = 0.90 and top@k = 3, so the improvement in matching performance is not statistically significant, but the matching validator will have a great impact on performance with lower similarity thresholds and higher top@k values.

\subsubsection{Matching Merger} 

\begin{figure}[htbp]
\centering
\subfloat{\includegraphics[width=0.333\columnwidth]{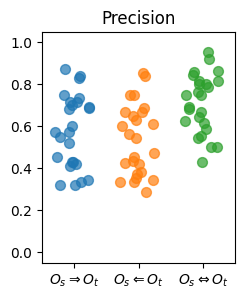}}
\subfloat{\includegraphics[width=0.333\columnwidth]{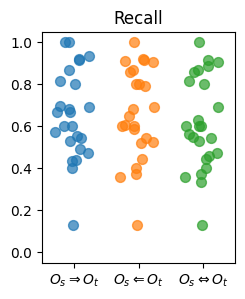}}
\subfloat{\includegraphics[width=0.333\columnwidth]{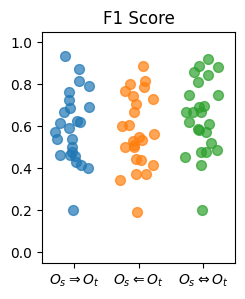}}
\caption{Comparison of $\mathbf{O_s \Rightarrow O_t}$, $\mathbf{O_s \Leftarrow O_t}$, and $\mathbf{O_s \Leftrightarrow O_t}$}
\label{fig: merge}
\end{figure}

We apply a merge function $O_s \Leftrightarrow O_t$ combining the results of $O_s \Rightarrow O_t$ and $O_s \Leftarrow O_t$ to improve the matching performance. Figure~\ref{fig: merge} shows the comparison of precision, recall, and F1 score in $O_s \Rightarrow O_t$, $O_s \Leftarrow O_t$, and $O_s \Leftrightarrow O_t$ in the three OAEI tracks we analysed. The merged matching results generally achieved a significant improvement in precision and F1 score, with a slight decrease in recall. The results are in line with the findings in RAG-Fusion~\cite{ragfusion}, where providing two different paths to perform the same matching task can reduce LLM hallucinations.

\subsection{Hyperparameter Settings}
\label{ablation: hyperparameter} 

\subsubsection{Similarity Threshold}

We test the similarity threshold $T\in[0.50, 0.55, 0.60,...,0.90, 0.95, 1.00]$ in the three OAEI tracks we analysed. The optimal similarity threshold appears to be $T\in[0.90, 0.95]$, balancing the trade-off between precision and recall, thus achieving a higher overall F1 score. If we consider the similarity threshold as the required confidence interval (CI) for candidates for equivalence matching, this setting reflects the convention of accepting a 5-10\% probability of observing values outside the estimation. The insensitivity of the results to the threshold value is more apparent in large-scale OM tasks, such as the OAEI Anatomy Track.

\subsubsection{Top@k}

We also test the top@k values $ k\in[1, 2, 3,..., 8, 9, 10]$ in the three OAEI tracks we analysed. We observe that $k=1$ and $k=2$ do not provide enough candidates for the LLM to select, while appropriate correspondences are rarely found where $k>5$. We recommend setting $k\in[3, 4, 5]$ to balance the computational complexity and precision of the results. Note that we deal with the tie-break case where multiple entities have the same RRF scores. In such cases, the total number of entities tested may be greater than $k$ because these equally-scored entities share the same ranking.

\subsubsection{How to choose Similarity Threshold and Top@k?}

Ontologies are context-dependent conceptual models that follow different conventions and restrictions to reflect different application-level requirements~\cite{ontology2021conceptual}. The hyperparameter settings can be adjusted for each specific OM task using reference mappings to achieve an optimal result. We observe that higher similarity thresholds and lower top@k values could result in high precision where most of the trivial mappings can be found, but some more obscure true mappings may be missed, lowering recall. On the other hand, lower similarity thresholds and higher top@k values could result in high recall, but the precision may be low as more false mappings are generated during the matching process. This indicates that the applied matching refinements (validator and merger) would be more powerful in settings with lower similarity thresholds and higher top@k values. We found in our extensive OAEI experiments that threshold $T\in[0.90, 0.95]$ and top@k $k\in[3, 4, 5]$ were optimal. In the real-world application of Agent-OM to a matching problem with no reference, we advise choosing $T$ and $k$ within these ranges.

\section{Discussion}
\label{sec: discussion}

Google DeepMind classifies AI autonomy into 6 levels~\cite{morris2024levels}. We believe that the potential of LLMs is not only as a consultant, a collaborator, or an expert to answer binary classification questions in OM tasks, but also as an agent to simulate human behaviour in performing OM tasks, including data preprocessing, data preparation, data analysis, and data validation. Higher autonomy reduces barriers to accessing LLMs in OM tasks.

\begin{enumerate}[wide, noitemsep, topsep=0pt, labelindent=0pt]

\item LLM-agent-based OM is more efficient than LLM-based OM. LLMs are computationally expensive. While LLM-based OM using binary classification questions has repetitive LLM prompting, LLM-agent-based OM leverages the vector database to store ontology retrieval results, reducing the financial cost of token consumption.

\item LLM-agent-based OM is also more effective than LLM-based OM. LLM-based OM is commonly observed to have limitations. Due to the nature of its large knowledge base, it is possible to discover positive correspondences but also to find unavoidable false mappings or missing true mappings. LLMs are zero-shot reasoners~\cite{kojima2022large}, but they are also few-shot learners~\cite{brown2020language}. Their capacity for reasoning depends on the richness of the information provided. With the assistance of autonomous agents for extensive planning, tools, and memory, LLM-agent-based OM can unlock the potential of LLMs and therefore feature the following advantages:

\begin{enumerate}[wide, noitemsep, topsep=0pt, labelindent=0pt]
\item Context Learning: LLMs have a large corpus of background knowledge. Given a context, LLMs can select relevant background knowledge and therefore perform better in lexical matching.
\item Transitive Reasoning: LLMs can reason on transitive relationships. They can also understand general and domain-specific scenarios and apply lexical validation when necessary.
\item Self Correction: LLMs have a strong capacity for self correction. Even given a wrong statement, LLMs have good judgement to automatically remove false mappings. For example, semantic matching can cause false mappings because it considers only the data structure and ignores the linguistic meaning of the entity. However, such a small piece of false information does not influence the correct truth that LLMs nevertheless learn.
\end{enumerate}

\end{enumerate} 

Despite the success of agent-powered LLMs for OM, there may be further opportunities for improvement as follows.

\begin{enumerate}[wide, noitemsep, topsep=0pt, labelindent=0pt]
\item A matching process could be more complex. Although CoT may simulate how humans plan and perform tasks, it is still an incomplete model of human thought. Human reasoning employs a more complex network of thoughts, as humans tend to try different isolated paths (i.e. ToT, tree of thoughts~\cite{yao2023tree,long2023large}), explore multiple paths (i.e. GoT, graph of thoughts~\cite{besta2024graph,yao2024got}), and backtrack, split, or merge to find the optimal solution to the problem. For example, people may use discovered mappings as input to the next iteration.
\item Prompt engineering is the key to instructing efficient LLM agents. These prompts are currently hand-crafted. For prompt-based tools, different LLMs may have varying default chat templates. Finding generic prompts across all LLMs is almost impossible. However, we provide the simplest standardised version of the prompts from our experiments. The prompts used in our system currently support mainstream LLMs, such as OpenAI GPT models, Authropic Claude models, Meta Llama 3, Alibaba Qwen 2, Google Gemma 2, and ChatGLM 4. For those models not included in the list, we also provide an interface to add new LLMs to our system, but it may require minor code customisation to fit the LLMs used. We expect that our system will support more models via open-source community efforts in the future. We seek automatic prompt engineering and will consider using soft prompts in future versions.
\item LLM hallucinations can be mitigated, but cannot be eliminated. The accuracy of the RAG remains an open question. Human-in-the-loop may remain necessary~\cite{ouyang2022training}. Advanced RAG techniques, such as including the explanatory context in the RAG process~\cite{contextual-claude}, are promising directions.
\item There is a trade-off between precision and recall. Strict rules could result in high precision where most of the trivial true mappings can be found, but some obscure true mappings may be missing. On the other hand, loose rules could result in a high recall score, but the precision score may become very low as more false mappings are generated during the matching process.
\item For LLMs used for OM, we find Moravec's paradox~\cite{moravec1988mind}: ``the hard problems are easy and the easy problems are hard''(\textit{p192})~\cite{pinker2003language}. Although Agent-OM performs well in complex and few-shot OM tasks, it is not outstanding on simple OM tasks. We will also consider integrating the LLM-based approach with traditional knowledge-based and ML-based approaches.
\end{enumerate}

\section{Limitations}
\label{sec: limitations}

\balance

\begin{enumerate}[wide, noitemsep, topsep=0pt, labelindent=0pt]
\item We evaluate only the TBox matching datasets that match classes, object properties, and datatype properties. ABox matching datasets (including individual data instances) are not considered due to privacy concerns in our targeted application domain. Additional data engineering (e.g. data de-identification and fuzzing) may be required to apply LLMs to ABox matching datasets to avoid personal and sensitive information exposure.
\item Due to the high cost of API calls for API-accessed commercial LLMs, experiments with the newest commercial models (e.g. OpenAI o1 and Anthropic claude-3-opus) are not included in this study. According to our findings in Section~\ref{ablation: llm}, we hypothesise that these models could achieve better performance in OM tasks.
\item There may be additional resource requirements for running open-source LLMs locally. The run time for API-accessed commercial LLMs is controlled by the LLM providers.
\end{enumerate}

\section{Future Work}
\label{sec: future work}

\begin{enumerate}[wide, noitemsep, topsep=0pt, labelindent=0pt]
\item Multimodal OM: We have packaged our system into several natural language-based commands. It could be integrated with advanced LLM functions to support multimodal input, such as ontology diagrams and online seminars. The richer information sources might improve OM performance.
\item Multilingual OM: Agent-OM supports ontologies in multiple languages. We tested it on the OAEI MultiFarm Track, an adapted conference dataset with ontologies translated into nine different languages. The results are not included here because there are few benchmarks available.
\item Small language models (SLMs) for OM: SLMs (e.g. gemma-2-2b) are useful in resource-constrained devices, but they have problematic tool interfaces at present.
\end{enumerate}

\section{Conclusion}
\label{sec: conclusion}

In this paper, we introduce a new design paradigm for OM systems. Agent-OM, an agent-powered LLM-based framework, is proposed and implemented with a proof-of-concept system. We compare our system with state-of-the-art OM systems to perform different types of OM tasks. The system has shown a powerful capability to perform OM tasks at different levels of complexity, leveraging the potential of using LLM agents for OM. We also discuss our observations on the advantages and current limitations of using LLMs and LLM agents for OM tasks.

Our work focuses on large pre-trained foundation models that are impossible to retrain and hard to fine-tune. Our approach yields good results on LLMs for OM tasks without changing the LLM model itself, but by utilising CoT, ICL/RAG, and prompt engineering techniques. It is a simple, lightweight, and natural language-driven approach with high scalability. Agent-OM is all you need. While OM has been studied for two decades or more, we are now at a point where the goal of 100\% accurate, fully-automated, and domain-independent OM seems to be within reach.

\begin{acks}

The authors thank the reviewers for providing insightful comments. The authors thank Sven Hertling for curating the Ontology Alignment Evaluation Initiative (OAEI) datasets. The authors thank the organisers of the Conference Track (Ondřej Zamazal, Jana Vataščinová, and Lu Zhou), the Anatomy Track (Mina Abd Nikooie Pour, Huanyu Li, Ying Li, and Patrick Lambrix), and the MSE Track (Engy Nasr and Martin Huschka), for helpful advice on reproducing the benchmarks in OAEI 2022 and OAEI 2023. The authors thank Jing Jiang from the Australian National University (ANU) for helpful advice on the verbalisation tool used in the paper. The authors thank Alice Richardson from the ANU Statistical Support Network for helpful advice on the statistical analysis used in the paper. The authors thank the Commonwealth Scientific and Industrial Research Organisation (CSIRO) for supporting this project.

Agent-OM did not participate in the OAEI 2022 and 2023 campaigns. According to the OAEI data policy (retrieved December 1, 2024), ``OAEI results and datasets, are publicly available, but subject to a use policy similar to \href{https://trec.nist.gov/results.html}{the one defined by NIST for TREC}. These rules apply to anyone using these data.'' Please find more details from the official website: \url{https://oaei.ontologymatching.org/doc/oaei-deontology.2.html}.
\end{acks}

\newpage
\bibliographystyle{ACM-Reference-Format}
\bibliography{qiang-bibliography-vldb}

\newpage
\appendix

\includepdf[pages=-]{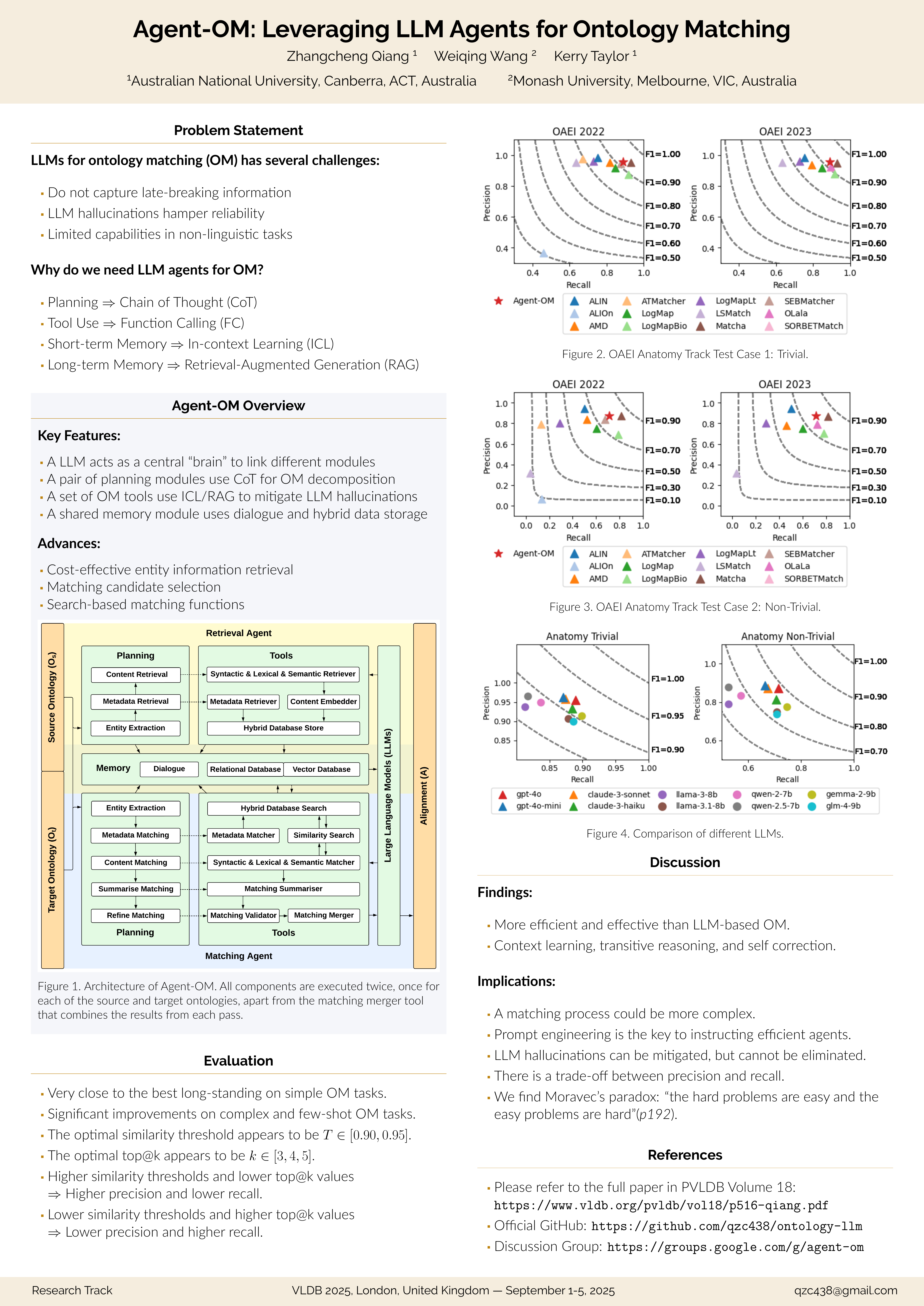}
\includepdf[pages=-]{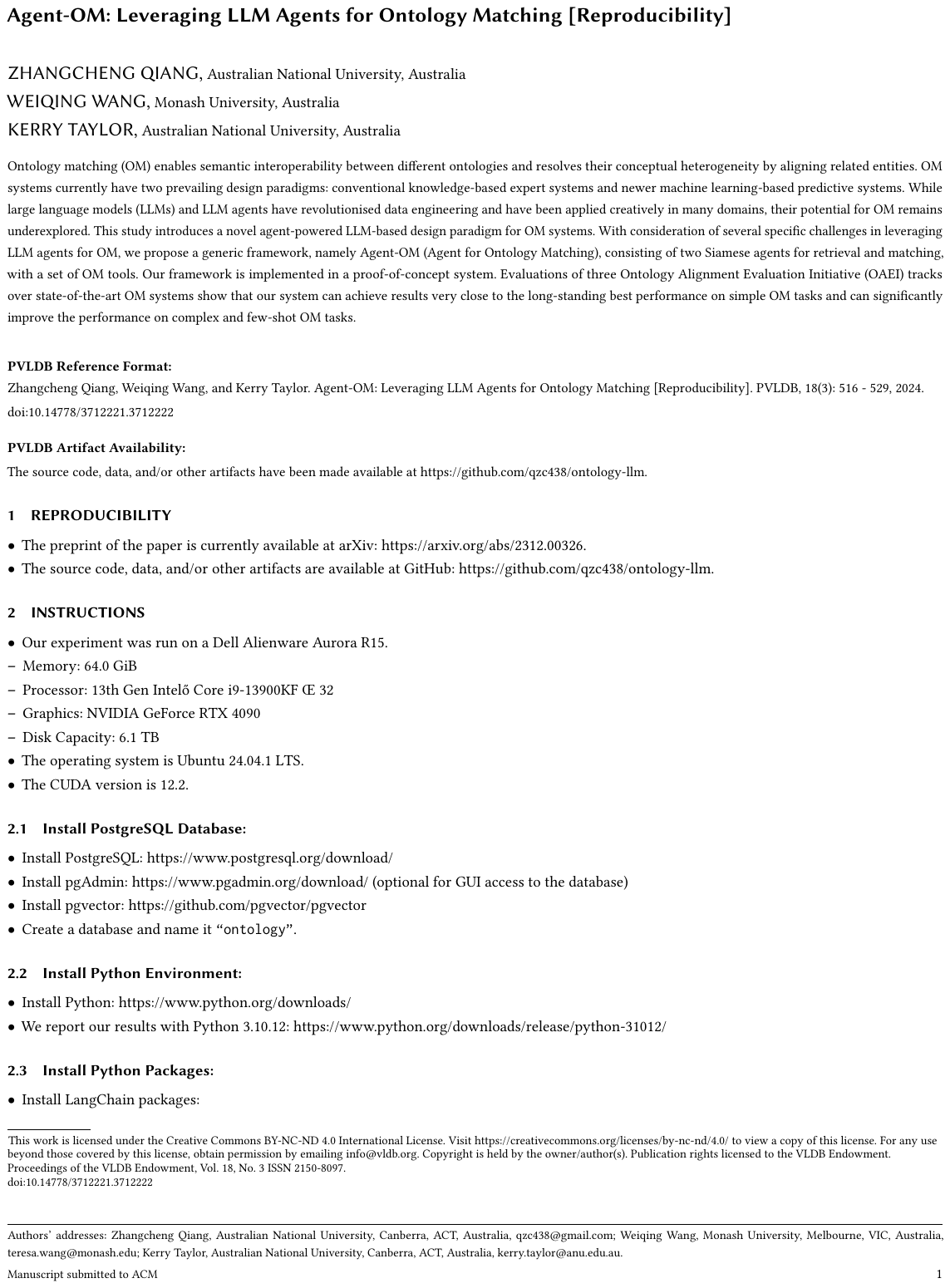}

\includepdf[pages=-]{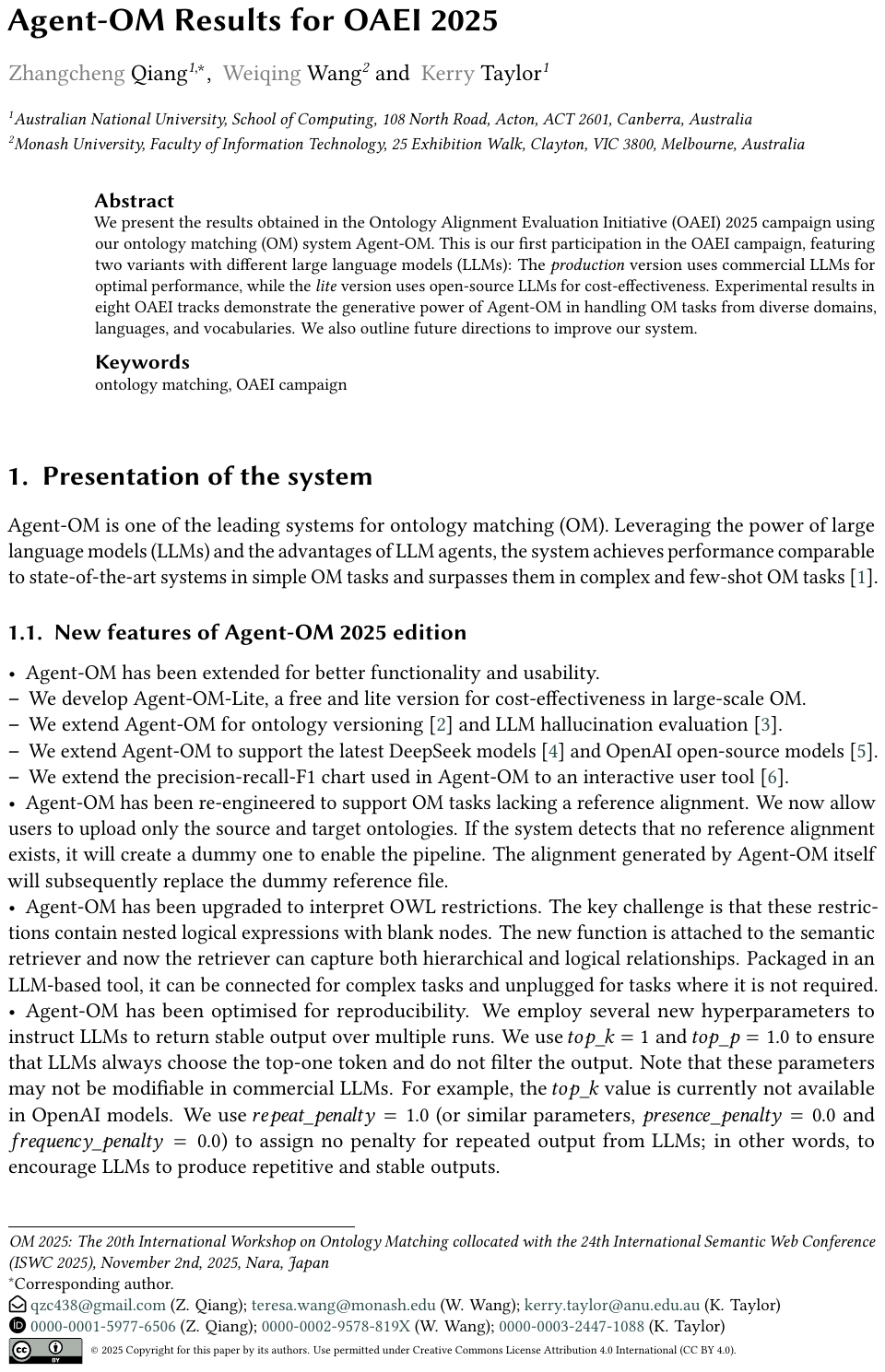}
\includepdf[pages=-]{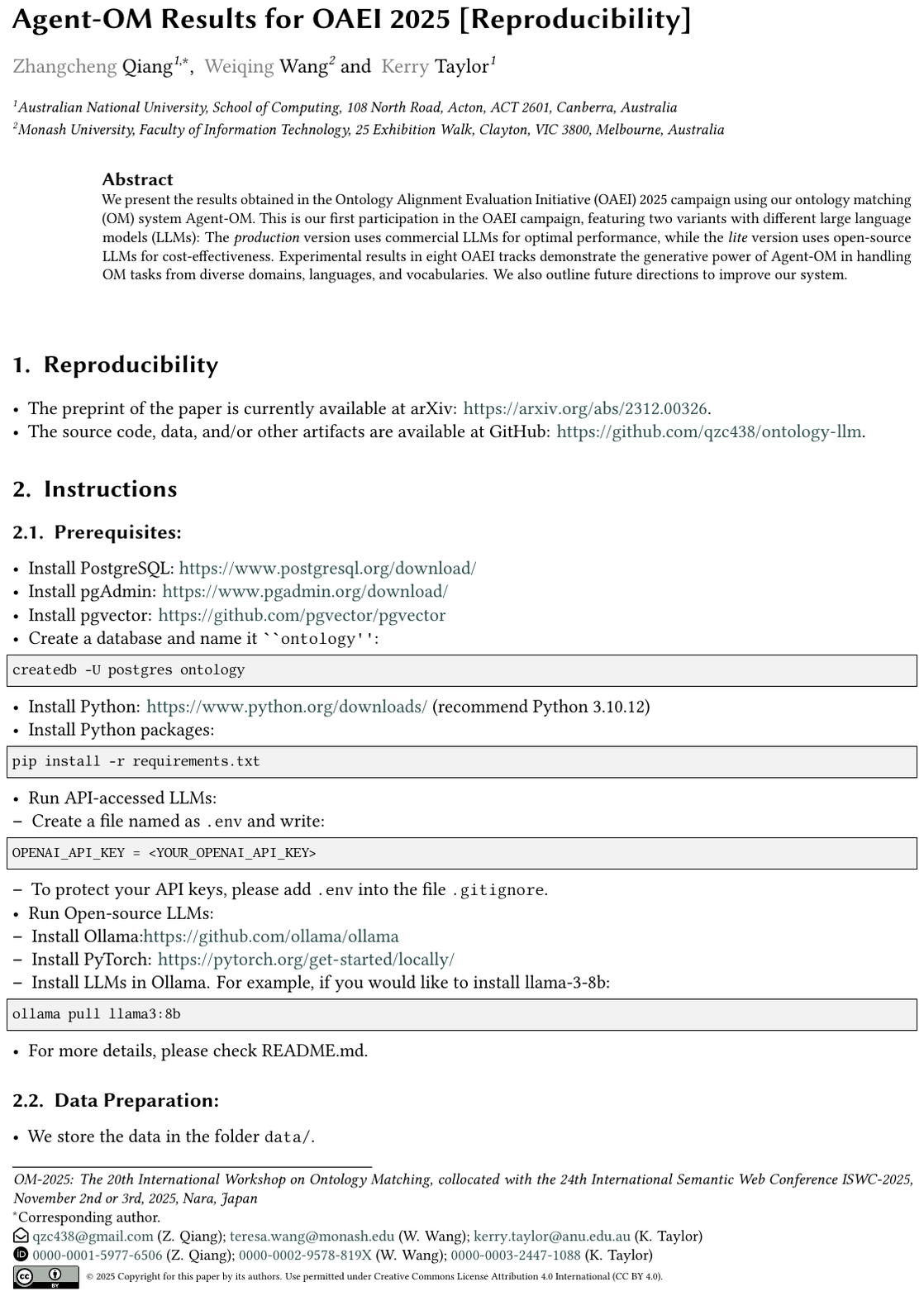}

\includepdf[pages=-]{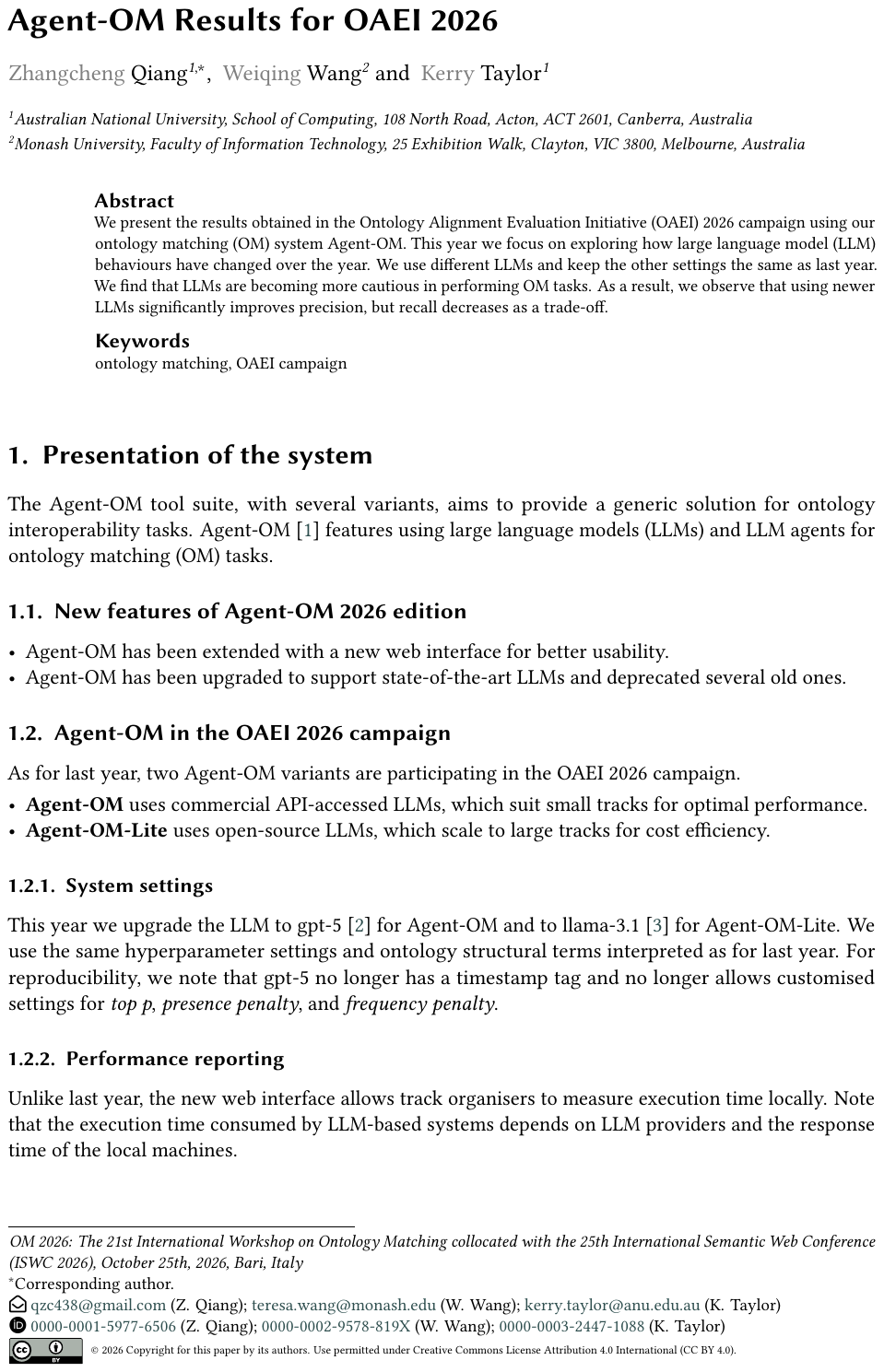}

\end{document}